\documentclass[runningheads]{llncs}

\usepackage{soul}
\usepackage{xcolor}
\definecolor{MyDarkRed}{rgb}{0.8,0.02,0.02}
\definecolor{MyRed}{rgb}{1.0,0.0,0.0}
\definecolor{airforceblue}{rgb}{0.36, 0.54, 0.66}

\renewcommand{\paragraph}[1]{\vspace{1.25mm}\noindent\textbf{#1}}

\usepackage{graphicx}
\usepackage{adjustbox}
\usepackage{multirow}
\usepackage{subfigure}
\usepackage{marvosym}
\begin{document}
\title{Audio-visual Saliency \\for Omnidirectional Videos}
\author{Yuxin Zhu \and
Xilei Zhu \and
Huiyu Duan \orcidID{0000-0002-6519-4067} \and
Jie Li \and
Kaiwei Zhang \and
Yucheng Zhu \and
Li Chen \and
Xiongkuo Min \and
Guangtao Zhai\textsuperscript{(\Letter)}
}
% \author{Anonymous Submission}
%
\authorrunning{Zhu et al.}
\institute{Institute of Image Communication and Network Engineering, \\Shanghai Jiao Tong University \\
\email{\{rye2000,~xilei\_zhu,~huiyuduan,~olivia605,~zhangkaiwei,~zyc420,~hilichen,\\~minxiongkuo,~zhaiguangtao\}@sjtu.edu.cn}\\
}
% \institute{Anonymous Institute}
%
\maketitle              % typeset the header of the contribution
\begin{abstract}
% The popularity of Virtual Reality(VR) has led to the generation of massive Omnidirectional Videos (ODVs).

Visual saliency prediction for omnidirectional videos (ODVs) has shown great significance and necessity for omnidirectional videos to help ODV coding, ODV transmission, ODV rendering, \textit{etc.}.
However, most studies only consider visual information for ODV saliency prediction while audio is rarely considered despite its significant influence on the viewing behavior of ODV.
This is mainly due to the lack of large-scale audio-visual ODV datasets and corresponding analysis.
Thus, in this paper, we first establish the largest audio-visual saliency dataset for omnidirectional videos (AVS-ODV), which
comprises the omnidirectional videos, audios, and corresponding captured eye-tracking data for three video sound modalities including mute, mono, and ambisonics.
Then we analyze the visual attention behavior of the observers under various omnidirectional audio modalities and visual scenes based on the AVS-ODV dataset.
Furthermore, we compare the performance of several state-of-the-art saliency prediction models on the AVS-ODV dataset and construct a new benchmark.
Our AVS-ODV datasets and the benchmark will be released to facilitate future research.

\vspace{-8pt}
\vspace{-8pt}
\end{abstract}
%
%
%
%\vspace{-20pt}
\section{Introduction}
\vspace{-6pt}
With the development of multimedia technology \cite{duan2023masked,wang2023aigciqa2023} and the popularity of Virtual Reality Head-mounted Displays (VR-HMDs) \cite{duan2018perceptual,duan2017ivqad}, more and more VR contents are generated.
Among these VR contents, Omnidirectional Videos (ODVs) can present $360^{\circ}$ real-world scenes, which provide more immersive visual experience.
Due to the great immersive experience, ODVs have been widely used in many application scenarios \cite{duan2023attentive,duan2022confusing,duan2022saliency,ren2023children}.
For example, in the advertising industry, YouTube has pioneered omnidirectional video advertising, which provides more detailed product information and has stronger user appeal; in the medical field, live ODV surgery can provide technical support for telemedicine training and promote the sharing of medical resources.

% \hdc{Users can explore ODVs in 3 Degrees of Freedom (3-DoF) under HMDs.}
% \hdr{In fact, when users consume omnidirectional videos, they do not focus on the entire scene but observe the field of view(FoV) of the omnidirectional video in the viewport through the HMD}{However, due to the limitation of the Field of View(FoV), at any moment, users can only see one viewport of an omnidirectional video}. 
Therefore, predicting the viewport movement, \textit{i.e.,} head movement \cite{tu2022end}, at the next moment can contribute to video encoding and transmission by reducing redundant information and improving efficiency.
Moreover, utilizing eye-tracking technology and foveated rendering can greatly enhance the visual experience \cite{duan2018perceptual,duan2019perceptual,duan2017ivqad,zhu2023perceptual}.
Thus, it is significant and necessary to study computing visual saliency of omnidirectinal videos and predicting the viewing behavior of users for a specific omnidirectional video.
% \hdr{We call the user's area of visual attention as the saliency area. If the saliency area can be well predicted, we can process omnidirectional videos targetedly during video encoding and video transmission, effectively reducing redundant information in data processing and improving the efficiency. In addition to guiding encoding and transmission, omnidirectional video saliency prediction has high application value in video rendering, object recognition, video editing, and video summarization. Therefore, it is necessary to predict saliency in omnidirectional videos, identify areas of interest for the human eye in the video, and predict users' viewing behavior for a specific omnidirectional video.}{}

Many saliency prediction studies for traditional images and videos have been conducted~\cite{6871397,duan2019visual,fang2020identifying,6857361,730558,yang2019predicting}. 
Moreover, some studies have also explored the problem of saliency prediction for omnidirectional images or videos \cite{ZHU201815,zhu2021viewing}.
However, as an important part of ODVs, omnidirectinal audios are rarely considered in omnidirectional saliency prediction studies, though it may strongly influence the human viewing behavior under VR-HMDs \cite{li2022sound}.
Therefore, it is necessary to conduct deeper research for saliency prediction in omnidirectional videos considering both visual and auditory dimensions. 

In this work, we establish a large-scale audio-visual saliency dataset for ODVs (AVS-ODV), analyze the impact of different audio modalities on ODV saliency, and conduct a comprehensive benchmark study. 
Specifically, we first capture and establish a large-scale omnidirectional video dataset with 8K resolution and omnidirectional sound.
Based on this video dataset, we collect eye-tracking data under a VR-HMD for three audio modalities including mute, mono, and first-order ambisonics (FOA), respectively.
The omnidirectional videos, audios, and corresponding eye-tracking data together constitute the established AVS-ODV dataset.
Then based on the constructed AVS-ODV dataset, we explore the visual attention mechanism of observers under omnidirectional audio-visual environments, analyze and summarize their viewing characteristics by comparing the differences of attention distribution under three different audio modalities.
Finally, we also compare the performance of several state-of-the-art saliency prediction models on the AVS-ODV dataset, and construct a comprehensive benchmark comparison to facilitate future studies.

Overall, the contributions of this paper are summarized as follows:
\vspace{-3pt}
\begin{itemize}
    \item To the best of our knowledge, our established AVS-ODV dataset is the largest audio-visual eye-tracking dataset for ODVs with the highest resolution.
    \item We conduct a comprehensive analysis based on the AVS-ODV dataset and explore the factors that significantly influence the human viewing behavior under ODVs.
    \item An extensive benchmark study is conducted based on the AVS-ODV dataset to facilitate future research.
\end{itemize}

\vspace{-18pt}
\section{Related Work}
\vspace{-6pt}
\subsection{Omnidirectional Video Saliency Dataset} 
\vspace{-6pt}
Table~\ref{dataset} provides an overview of several widely used omnidirectional video saliency datasets.
For subjective eye-tracking experiments of ODVs, both head movement (HM) data and eye movement (EM) data are usually recorded, where HM means the viewport direction of head and EM reveals the specific fixation location \cite{7840720,Corbillon2017360DegreeVH,fremerey2018avtrack360,lo2017360,8463418,xu2017subjective}.
The distributions of HM and EM can be very different, since even though the head is stationary, viewers can still observe omnidirectional videos through eye movements.
Early studies only collect HM data to construct omnidirectional video saliency datasets \cite{7840720,Corbillon2017360DegreeVH,fremerey2018avtrack360,lo2017360,8463418,xu2017subjective}.
Recently, with the improvement of eye tracking technology and processing of algorithms, many studies collect both HM and EM information in subjective eye-tracking experiments \cite{David2018ADO,8418756,8578657,zhang2018saliency}.
%In the saliency experiment, viewport movement recording methods include head movement (HM) and eye movement (EM), where HM means the position of the subject's gaze on the  omnidirectional images and videos, while EM means the position of the subject's fixation within the gaze. 
%Since when the head is stationary, the subject can still view omnidirectional videos by moving the eyeballs, the distributions of HM and EM may differ. 
%In earlier datasets, saliency benchmarking was primarily based on HM data\cite{7840720,Corbillon2017360DegreeVH,fremerey2018avtrack360,lo2017360,8463418,xu2017subjective}. 
%With the improvement of eye tracking technology and processing of algorithms, more datasets have integrated both HM and EM information in their studies\cite{David2018ADO,8418756,zhang2018saliency,8578657}.

\begin{table}[!t]
\caption{An overview of omnidirectional video saliency datasets. ``Mute" means mute audio, ``mono" represents one channel audio and ``ambisonics" indicates first-order ambisonics audio. HM and EM represent head movement and eye movement respectively.}\label{dataset}
\vspace{-5pt}
\centering
\begin{adjustbox}{width=\textwidth}
\begin{tabular}{|c|c|c|c|c|c|c|}
\hline
Dataset &  Video Num & Audios & resolution &  Duration(s) & Data Type & Participants\\
\hline
Bao \textit{et al.}\cite{7840720}                        & 16   & Mute            & 2K,4K                    & 30          & HM     & 153       \\ 
Lo \textit{et al.}\cite{lo2017360}                       & 10   & Mute            & 4K                       & 60          & HM     & 50        \\ 
Corbillon \textit{et al.}\cite{Corbillon2017360DegreeVH} & 5    & Mute            & 4K                & 70         & HM     & 59        \\
Xu \textit{et al.}\cite{xu2017subjective}                & 48   & Mute            & 3K$\sim$8K & 20$\sim$60 & HM     & 40        \\ 
AVTrack\cite{fremerey2018avtrack360}       & 20   & Mute            & 4K                       & 30          & HM     & 48        \\ 
Ozcinar \textit{et al.}\cite{8463418}                    & 6    & Mute            & 4K, 8K             & 10          & HM    & 17        \\ 
Salient360\cite{David2018ADO}              & 19   & Mute            & 4K                & 20          & HM+EM  & 57        \\ 
PVS-HM\cite{8418756}                       & 76   & Mute            & 3K$\sim$8K               & 10$\sim$80 & HM+EM  & 58        \\ 
Zhang \textit{et al.}\cite{zhang2018saliency}            & 104  & Mute            & 4K                      & 20$\sim$60 & HM+EM  & 27        \\ 
VR-EyeTracking\cite{8578657}               & 208  & Mono          & $\geq$4K          & 20$\sim$60 & HM+EM  & 31        \\ 
Chao \textit{et al.}\cite{9105956}                        & 12   & Mute/Mono/Ambisonics   & 4K                & 25          & HM     & 3$\times$15 \\ 
Li \textit{et al.}\cite{li2022sound}                         & 46   & Mute/Mono       & 4K                       & 15          & HM+EM  & 2$\times$15 \\ 
\textbf{AVS-ODV (ours)} & \textbf{162} & \textbf{Mute/Mono/Ambisonics} & \textbf{8K} & \textbf{15} & \textbf{HM+EM} & \textbf{3$\times$20} \\
% Robotham \textit{et al.}\cite{9900893}               & 12   & Ambisonics & 7680$\times$3840                & about 60       & none      & none        \\
\hline
\end{tabular}
\end{adjustbox}\vspace{-18pt}
\end{table}

\vspace{-10pt}
\subsection{Saliency Prediction Models}
\vspace{-8pt}
\paragraph{\textit{Traditional saliency prediction.}}
%Itti \textit{et al.}'s \cite{730558} model is considered the pioneering work of traditional saliency detection. This model first extracts brightness, color, and orientation features of images at multiple scales and fuses them into a saliency map. 
Itti \textit{et al.} \cite{730558} proposed a computational efficient saliency model, which first extracts brightness, color, and orientation features of images at multiple scales and fuses them into a saliency map. 
Cheng \textit{et al.} \cite{6871397} proposed an image saliency prediction algorithm based on histogram contrast (HC), which first performs color quantization in the RGB color space, and then calculates distance using color statistical histograms in the Lab color space.
With the success of saliency prediction on images, many studies have also researched the problem of the saliency prediction on videos. Fang \textit{et al.} \cite{6857361} proposed a Spatio-Temporal Uncertainty Weighting (STUW) algorithm to detect visual saliency in video signals by combining spatial and temporal information with statistical uncertainty measures. %Wang \textit{et al.} \cite{7837719} mixed low-level features from psychophysical stimuli into a unified geodesic distance-based framework, resulting in reliable and temporally consistent saliency prediction maps at the superpixel level.

\paragraph{\textit{Omnidirectional video saliency prediction.}}
With the popularity of VR-HMDs, many studies have also explored predicting visual saliency for omnidirectional images and videos.
%There have been studies on saliency prediction for omnidirectial images and videos using handcrafted features. 
Zhu \textit{et al.} \cite{ZHU201815} proposed a saliency prediction method for head and eye movements in omnidirectional scenes by utilizing four different low-level features and high-level features. With the rapid development of deep learning, many omnidirectional video saliency prediction models have been proposed based on deep learning. Cheng \textit{et al.} \cite{8578252} introduced a simple and effective Cube Padding in 360$^{\circ}$ Videos (CP360) technique and proposed a spatial temporal saliency prediction method based on deep neural networks. 
Zhang \textit{et al.} \cite{zhang2018saliency} defined a spherical convolutional neural network that shares kernels over all patches on the sphere and instantiates a spherical U-Net structure for frame-by-frame saliency prediction. 
Xu \textit{et al.} \cite{8418756} proposed a DRL-Based Head Movement Prediction (DHP) method based on deep reinforcement learning to model attention on omnidirectional videos.

% \textbf{\textit{Sound Signals Researches}}  

% In early research, video saliency prediction mainly used visual single-modal information, ignoring auditory information. However, more and more studies have shown that sound signals have a significant impact on visual attention. Chao \textit{et al.} \cite{9105956} demonstrated the importance of spatial audio in ODV through experiments. They found that when watching muted ODVs, users' visual attention was widely dispersed, while attention was concentrated on salient areas when watching monaural and binaural ODVs. Salient audio clues (such as human voices and sirens) and visual clues (such as faces and moving objects) had a greater impact on visual attention. Li \textit{et al.} \cite{9897737} verified that the presence of sound can attract viewers' attention to the sound source and make the distribution of visual attention more focused, especially when there are multiple visual salient objects with only one sound source. 
\paragraph{\textit{Omnidirectional video audio-visual saliency prediction.}}
As an important part of omnidirectional videos, the influence of omnidirectional audios on visual attention has rarely been studied.
%However, relatively few audio-visual saliency prediction models exist for omnidirectional videos. 
Chao et al. \cite{9301766} extended the DAVE \cite{tavakoli2019dave} network and proposed an audio-visual saliency prediction network for omnidirectional videos based on spatial audio (AVS360). 
%In terms of vision, they applied CP \cite{8578252} to the convolutional and pooling layers of the 3D ResNet \cite{8578783} to reduce geometric distortion caused by projection in visual features. They combined the extracted CP format local spatiotemporal visual features with the ERP format global spatiotemporal visual features through average pooling layers. In terms of auditory processing, they calculated the audio energy distribution (Audio Energy Map or AEM) by direction from the four channels of spatial audio. After combining spatiotemporal audio-visual clues, the extracted features were input into the decoder network to obtain the final audio-visual saliency map. 
Cokelek et al. \cite{9511406} proposed an unsupervised salient spatial sound localization method (SSSL), and applied an extended Mel-Cepstrum-Based Spectral Residual Saliency Detection Model (MCSR) to each channel of FOA audio input to obtain a 3D vector representation of the audio direction in omnidirectional space. 
%They obtained the audio saliency map through weighted fusion with existing saliency models such as CP360 \cite{8578252} and STAViS \cite{Tsiami_2020_CVPR}, and found that the performance of the model was greatly improved.

\vspace{-8pt}

\begin{table}[!t]
\vspace{5pt}
\caption{
%Statistics on the number of ODVs by category
An overview of the count statistics for each category in our AVS-ODV dataset.
}\label{tb:dataset statistics}
\centering
\setlength{\tabcolsep}{6pt}
\vspace{-5pt}
\scalebox{0.8}{
\begin{tabular}{|c|c|c|c|c|}
\hline
~Type~ & ~Scene~ & ~Sound Source~ & ~Number~ & ~Number~ \\ \hline
\multirow{4}{*}{1}       & \multirow{2}{*}{Indoor}        & Moving       & 12   & \multirow{4}{*}{51}       \\ \cline{3-4}
                         &                           & Stationary       & 22   &                           \\ \cline{2-4}
                         & \multirow{2}{*}{Outdoor}        & Moving       & 12   &                           \\ \cline{3-4}
                         &                           & Stationary       & 5    &                           \\ \hline
\multirow{4}{*}{2}       & \multirow{2}{*}{Indoor}        & Moving       & 3    & \multirow{4}{*}{58}       \\ \cline{3-4}
                         &                           & Stationary       & 31   &                           \\ \cline{2-4}
                         & \multirow{2}{*}{Outdoor}        & Moving       & 5    &                           \\ \cline{3-4}
                         &                           & Stationary       & 19   &                           \\ \hline
\multirow{4}{*}{3}       & \multirow{2}{*}{Indoor}        & Moving       & 6    & \multirow{4}{*}{53}       \\ \cline{3-4}
                         &                           & Stationary       & 28   &                           \\ \cline{2-4}
                         & \multirow{2}{*}{Outdoor}        & Moving       & 15   &                           \\ \cline{3-4}
                         &                           & Stationary       & 4    &                           \\ \hline
\end{tabular}}\vspace{-13pt}
\end{table}

\vspace{-3pt}
\section{AVS-ODV Dataset}
\vspace{-6pt}
%%%%%%%%%%%%%%%XXX考虑一下和其他数据集详细对比XXX%%%%%%%%%%%%%%%%%%%%%%
%To the best of our knowledge, our proposed dataset is the first 8K saliency datasets with ambisonics. It also features the inclusion of user's visual attention information in mute, mono and ambisonics modalities.
In this section, we introduce the detailed process of constructing the AVS-ODV dataset including omnidirectional video collection and subjective eye-tracking experiment.
\vspace{-10pt}
\subsection{Video Stimuli}
\vspace{-4pt}
We first captured 162 different ODVs from different scenes with a professional VR camera Insta360 Pro2 \cite{insta360}.
%Our AVS-ODV dataset consists of 162 different ODVs taken with professional VR camera Insta360 Pro2. 
Each video has a resolution of 8K (7680$\times$3840) in equirectangular projection (ERP) format with a frame rate of 29.97 fps.
%, a duration of 15 seconds, and the frame rate is 29.97 fps. 
Then all raw videos were clipped to a duration of 15s for the convenience of data analysis and algorithm design.
The ODVs contain FOA with an audio sampling rate of 48,000 Hz. 
Four audio channels are arranged in the sequence of ``WYZX" (center, left-right, up-down, front-back).

\begin{figure}[!t]
\subfigure{
\rotatebox{90}{\scriptsize{\textbf{~Type 1}\label{predict:stavis2}}}
\includegraphics[width = .21\textwidth]{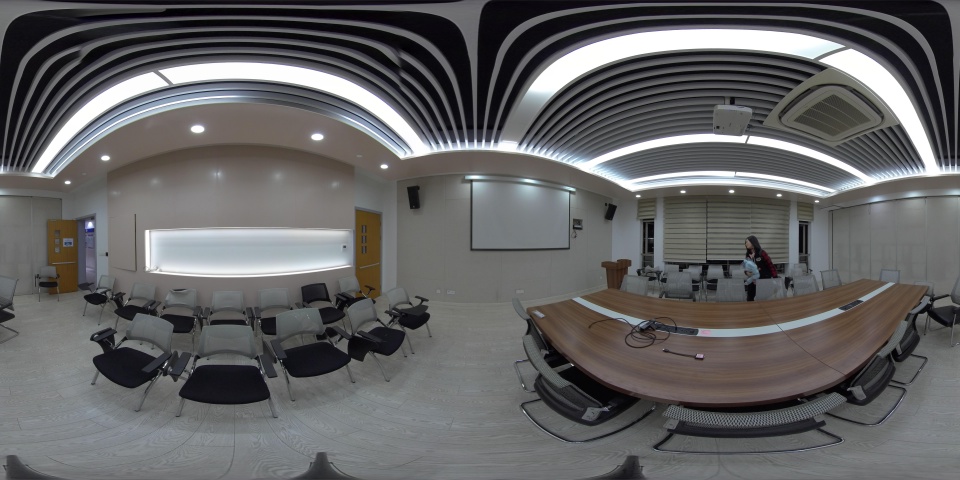}\hspace{-0.85mm}
\includegraphics[width = .21\textwidth]{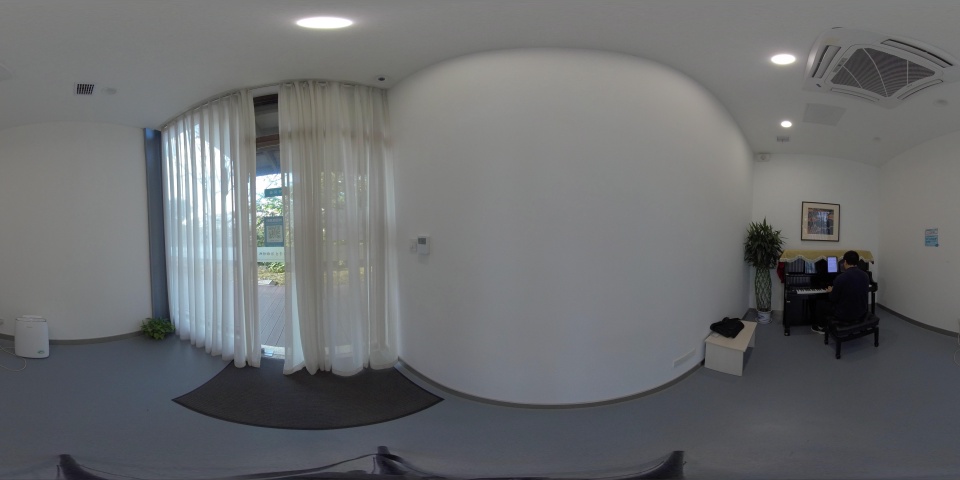}\hspace{-0.85mm}
\includegraphics[width = .21\textwidth]{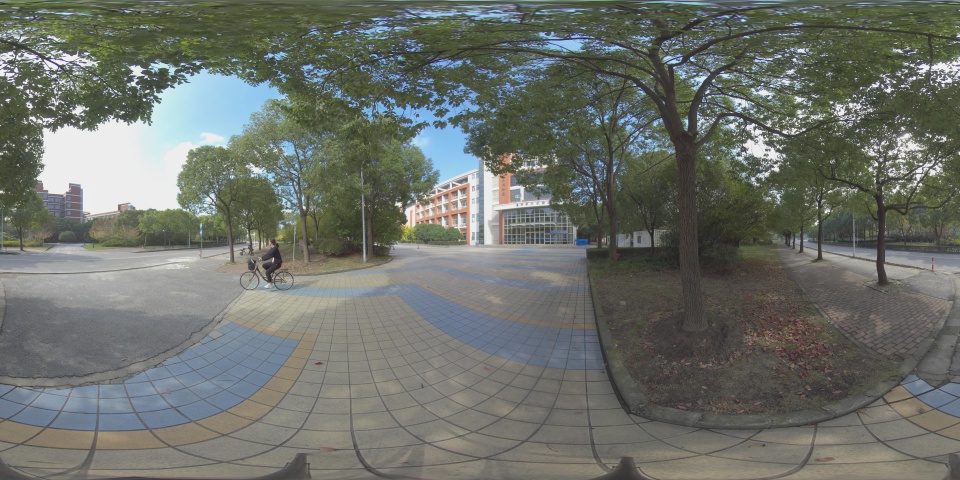}\hspace{-0.85mm}
\includegraphics[width = .21\textwidth]{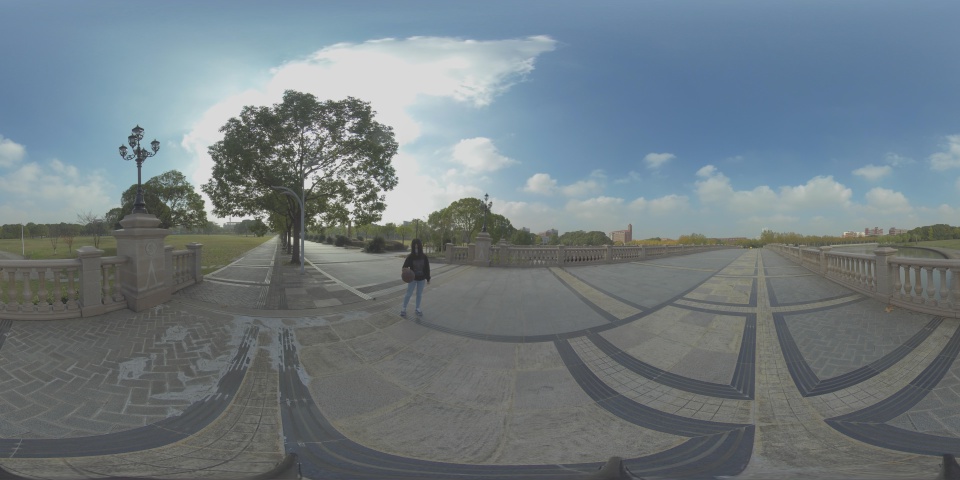}\hspace{-0.85mm}
}\vspace{-1mm}
\subfigure{
\rotatebox{90}{\scriptsize{\textbf{~Type 2}\label{sample-frames:stavis3}}}
\includegraphics[width = .21\textwidth]{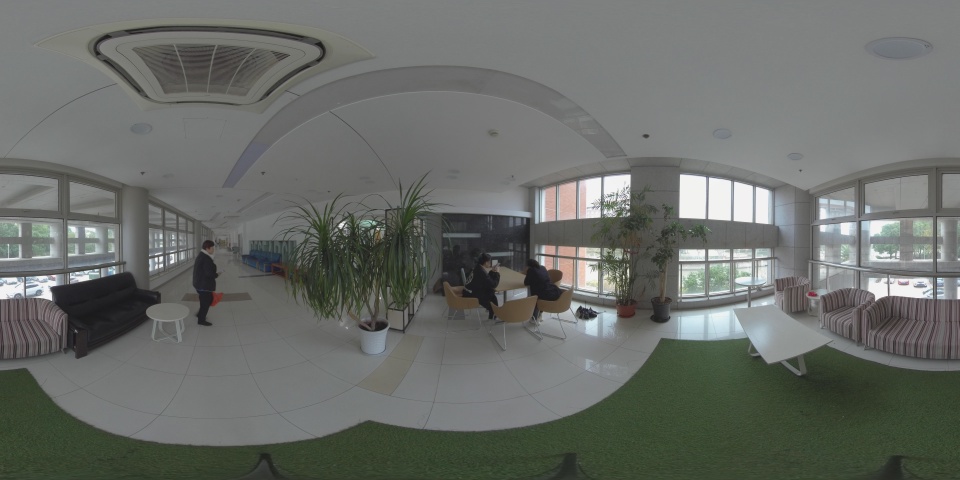}\hspace{-0.85mm}
\includegraphics[width = .21\textwidth]{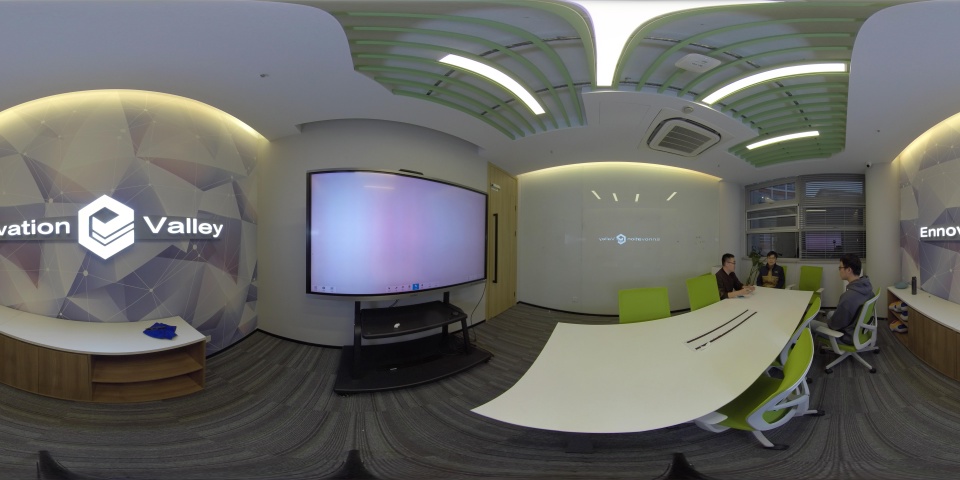}\hspace{-0.85mm}
\includegraphics[width = .21\textwidth]{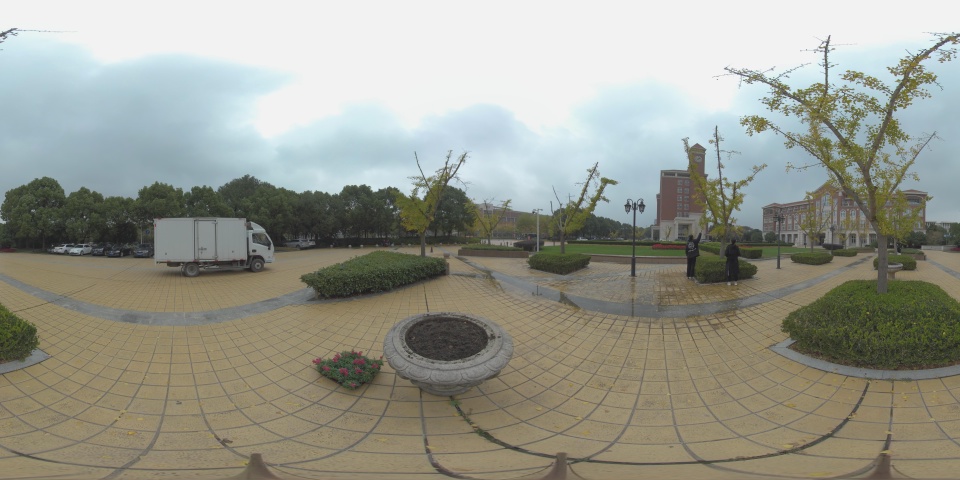}\hspace{-0.85mm}
\includegraphics[width = .21\textwidth]{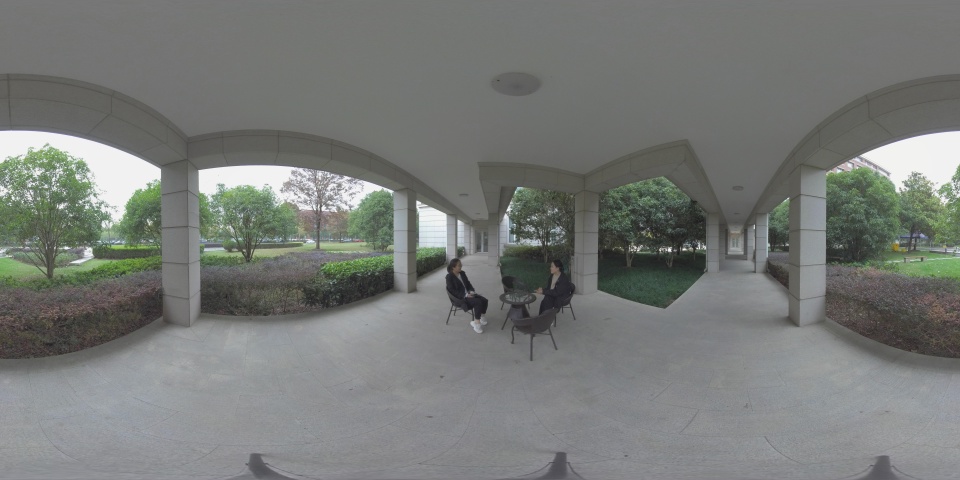}\hspace{-0.85mm}
}\vspace{-1mm}
\setcounter{subfigure}{0}
\subfigure[I \& M]{
\rotatebox{90}{\scriptsize{\textbf{~Type 3}\label{sample-frames:avs}}}
\includegraphics[width = .21\textwidth]{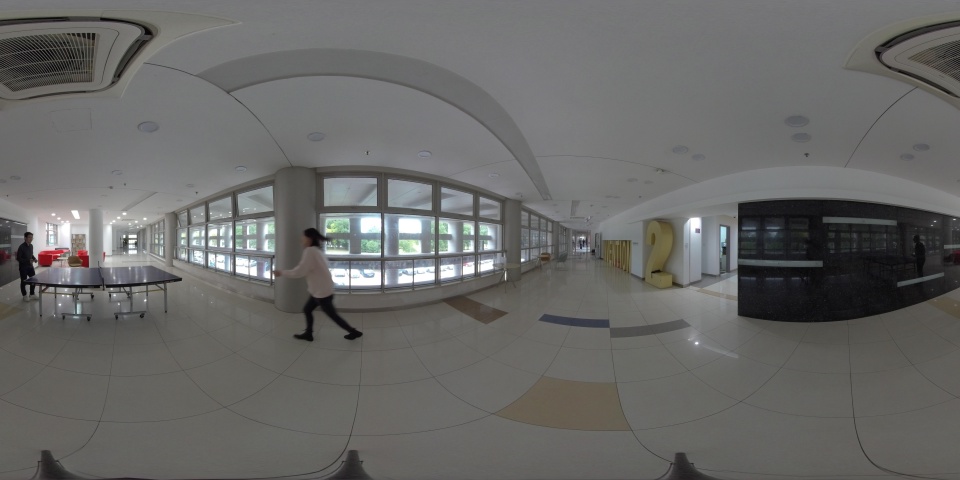}\hspace{-0.85mm}}
\subfigure[I \& S]{\includegraphics[width = .21\textwidth]{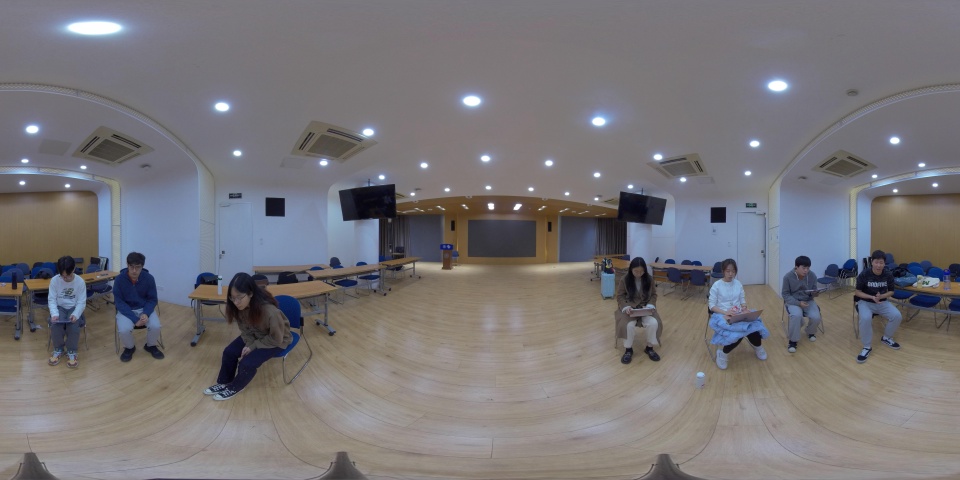}\hspace{-0.85mm}}
\subfigure[O \& M]{\includegraphics[width = .21\textwidth]{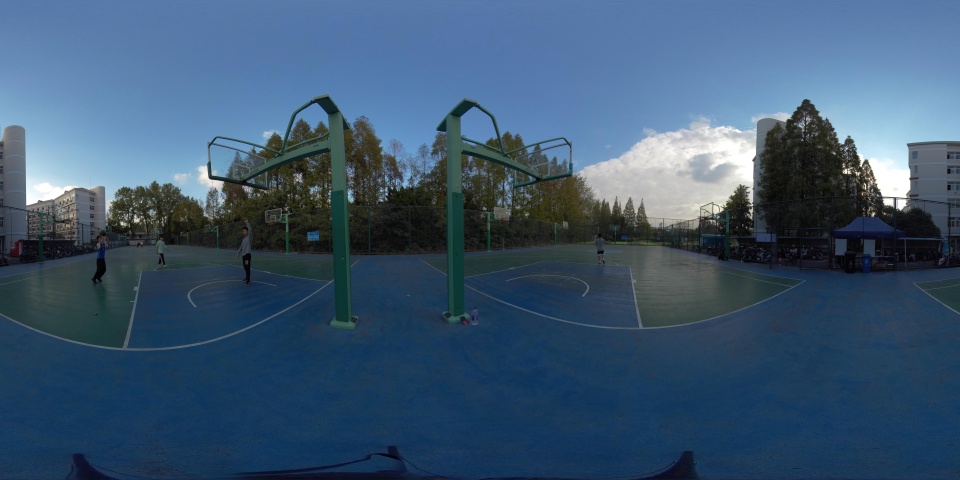}\hspace{-0.85mm}}
\subfigure[O \& S]{\includegraphics[width = .21\textwidth]{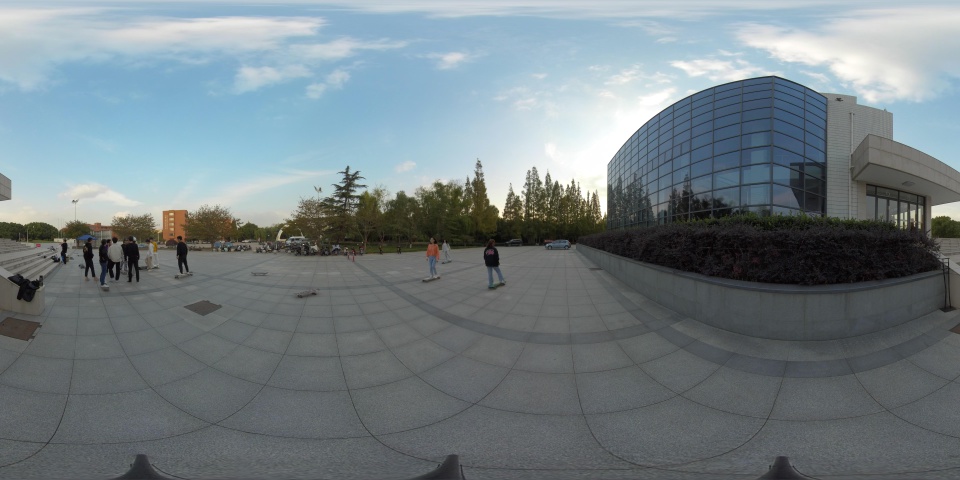}\hspace{-0.85mm}}\vspace{-10pt}
\caption{Examples of different categories (I: indoor, O: outdoor, M: moving, S:staionary).} \label{fig:sample video frames}
\vspace{-12pt}\end{figure}

%Inspired by the conclusions of Li \textit{et al.} \cite{li2022sound} , 
The collected 162 videos can be divided into three categories based on the number of salient objects and sound sources: (1) only one salient object and the sound source is that salient object, (2) multiple salient objects and the sound source is one of them, (3) multiple salient objects and multiple sound sources.
The salient objects mainly refer to humans or moving objects. 
Moreover, our AVS-ODV also contains various scenes including indoor scenes and outdoor scenes, and various types of sound sources including moving and stationary sound sources.
The detailed information of the ODV categories in our established AVS-ODV is demonstrated in Table~\ref{tb:dataset statistics}, and the analysis of the scene category on omnidirectinal visual attention can be found in Section \ref{sec:gaze_analysis}.
%We also take the scene features (outdoor or indoor) and sound source characteristics (moving or stationary) into account to further investigate their influence on visual attention. 
%We name the videos according to their categories, and the statistics of the number of videos in each category are shown in Table~\ref{tb:dataset statistics}. 
Fig.~\ref{fig:sample video frames} shows several examples of different ODV categories in our dataset.

\begin{figure}[!t]
\centering
\includegraphics[width=0.7\textwidth]{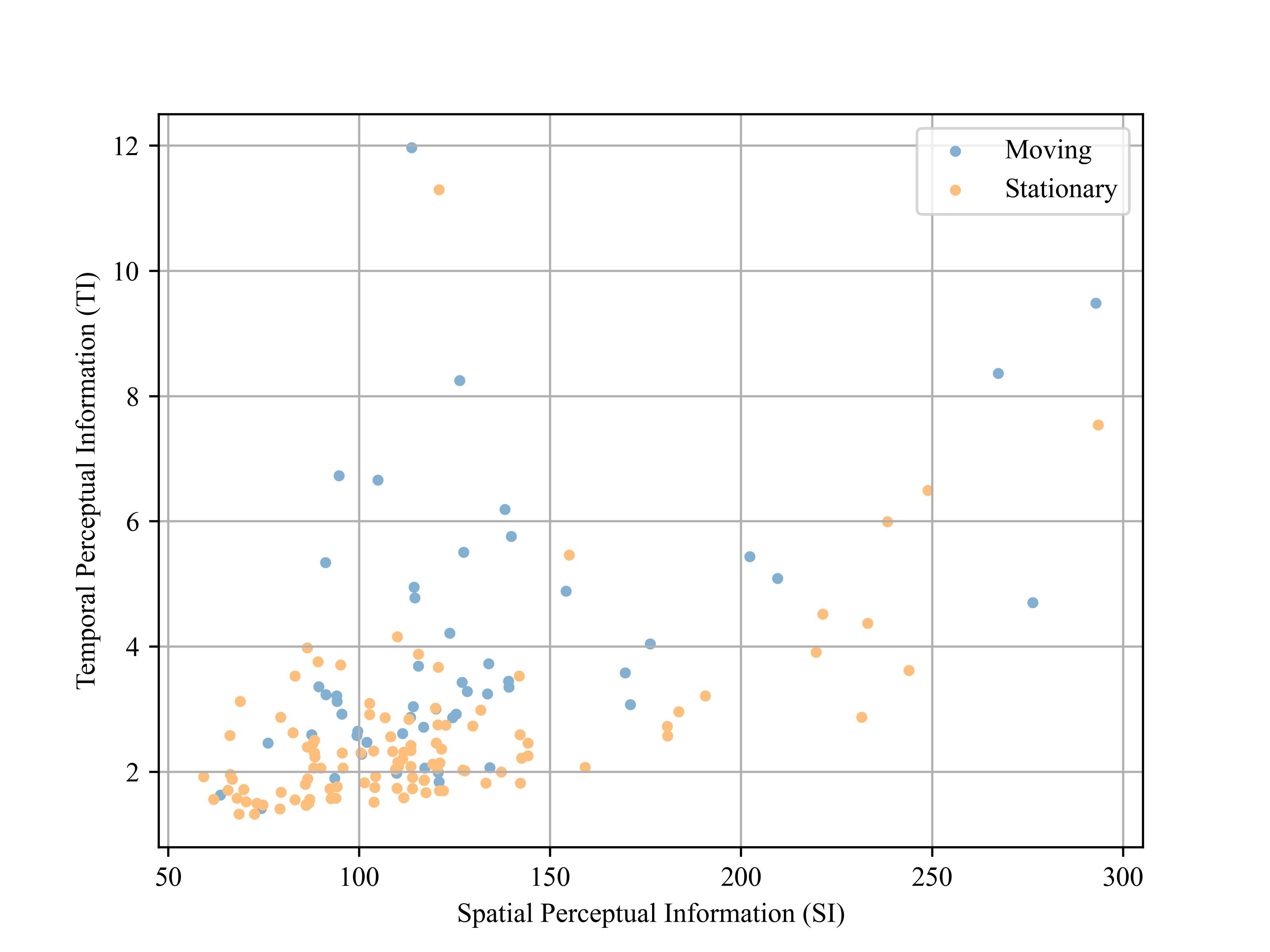}\vspace{-10pt}
\caption{SI and TI distributions of the ODVs in our dataset.} \vspace{-12pt}\label{fig:siti}
\end{figure}

SITI-tools %\cite{siti} 
can calculate spatial perceptual information (SI) and temporal perceptual information (TI) and is often used to check whether the materials used cover an appropriate range of spatial and temporal complexity. To reduce the impact of ERP distortion on the calculation of the video representation, we use the cubemap projection (CP) of the ODVs to calculate the corresponding SI and TI, and the final SI and TI score is calculated by the mean of the six cubic surfaces. From Fig.~\ref{fig:siti}, it can be observed that the videos have appropriate spatial and temporal complexity. Videos containing moving sound sources have higher temporal complexity in general, except for a very few videos with stationary sound sources that have high temporal complexity due to the large number of moving silent salient objects.

\vspace{-8pt}
\subsection{Subjective Experiment}
\vspace{-2pt}
\subsubsection{Apparatus.}
We used HTC Vive Pro Eye to play ODVs and record eye-movement data.
HTC Vive Pro Eye is an excellent VR-HMD with a bulit-in eye-tracker.
We developed a software based on Unity to conduct the subjective eye-tracking experiment.
Since the video codec that comes with Unity cannot meet the demand of smoothly playing 8K video, we applied the AVPro Media Player plugin to play the ODVs in our AVS-ODV smoothly.
%HTC Vive Pro Eye , a head-mounted display (HMD) with bulit-in eye-tracking devices, was used to play ODVs and record eye-movement. 
%For subjective experimentation, we used a software framework based on Unity as testbed. 
%As Unity's own video codec cannot meet the demand for smooth playback of 8K video, AVPro Media Player Ultra Edition plugin from Unity Asset Store was applied. 
The ambisonics was played using the Facebook Audio 360 audio tool to decode the audio and output it to the VR-HMD.

\vspace{-13pt}
\subsubsection{Subjects.}
In this work, we consider three audio modalities, including mute, mono and ambisonics, to explore the influence of audio on the visual attention on ODVs.
A total of 60 subjects were recruited. 
Eye movements were collected from 20 subjects for each audio modality, and one subject only see a source ODV once since viewing twice or more times may affect data reliability. 
We recorded the age and gender of the subjects, as shown in Table~\ref{tb:subject-info}. 
All subjects had normal or corrected-to-normal eyesight and hearing.

\begin{table}[!t]%%%%%%%%%%%%%%%%%%%%%%%%%%%%%%%有时间优化一下表格%%%%%%%%%%
\caption{Subjects' information.}\label{tb:subject-info}\vspace{-5pt}
\centering
\begin{tabular}{|c|c|c|}
\hline
~Modality~ & ~Age(range/mean/std)~ & ~Num(male:female)~ \\ \hline
mute       & 20-27/22.3/1.89     & 8:12             \\ 
mono       & 20-27/22.1/1.71     & 8:12             \\
ambisonics & 20-25/21.9/1.84     & 10:10            \\ \hline
\end{tabular}\vspace{-6pt}
\end{table}

\begin{figure}[!t]
\centering
\subfigure[Mute]{\includegraphics[width = .25\textwidth]{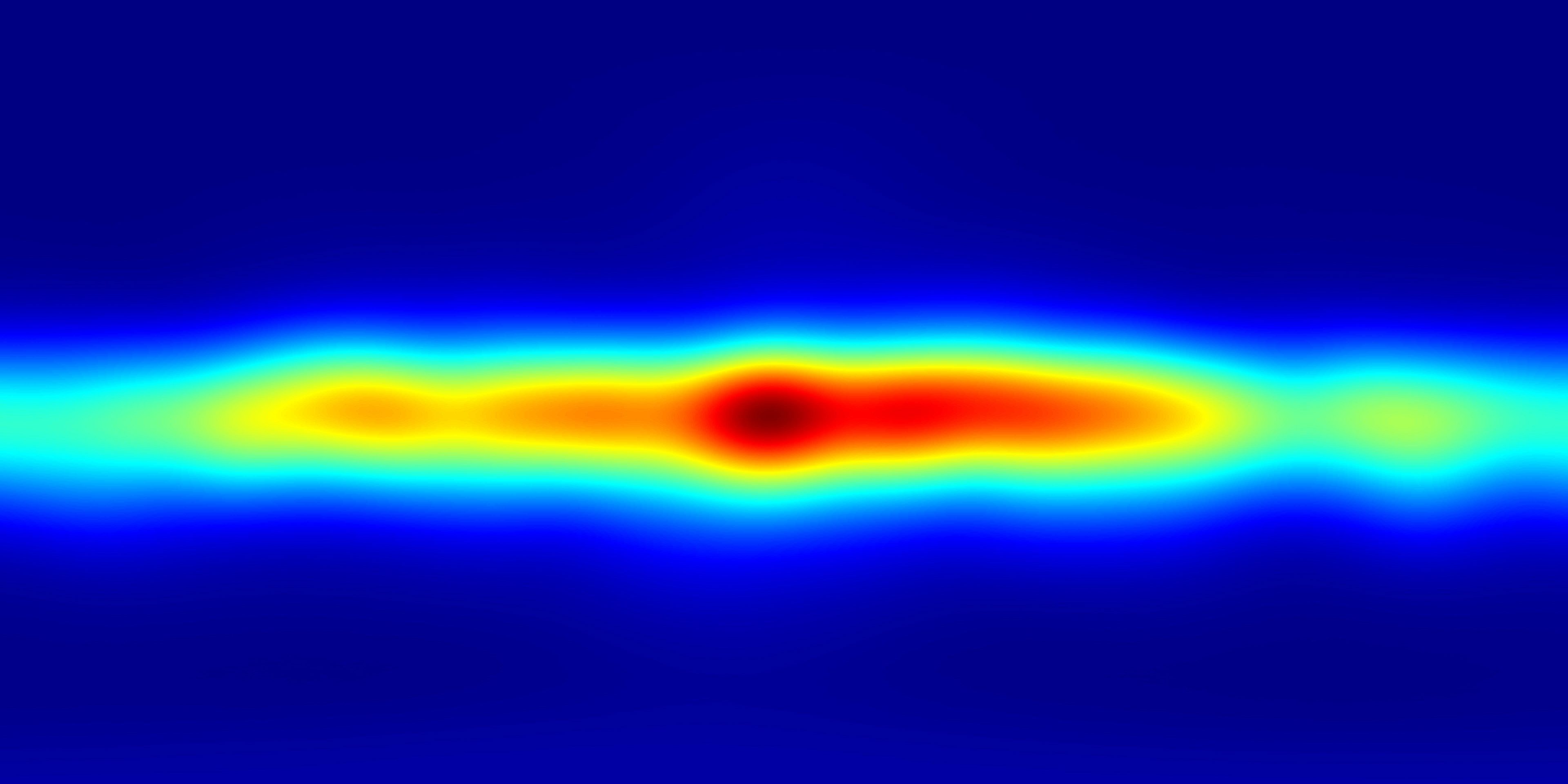}}
\subfigure[Mono]{\includegraphics[width = .25\textwidth]{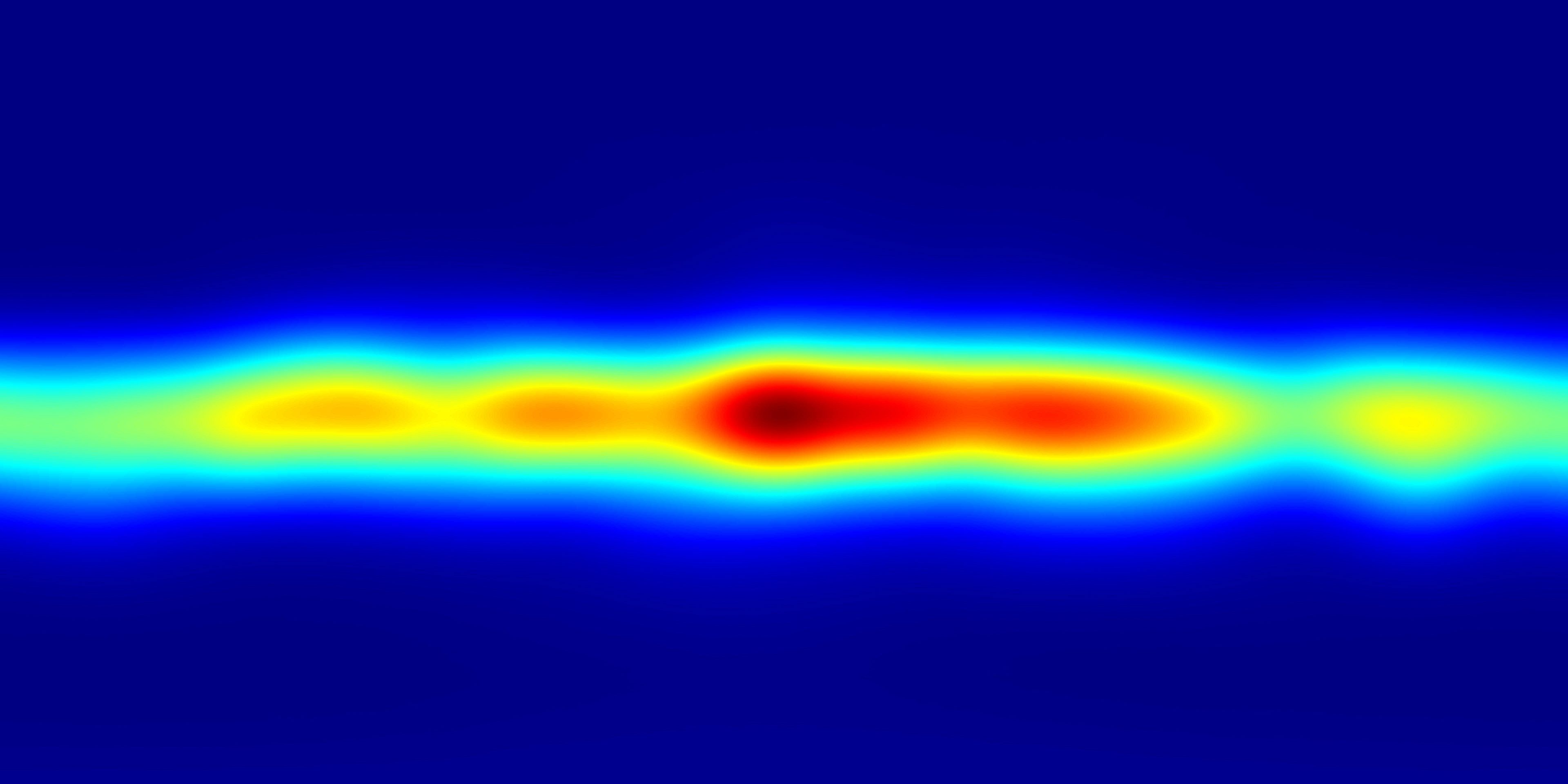}}
\subfigure[Ambisonics]{\includegraphics[width = .25\textwidth]{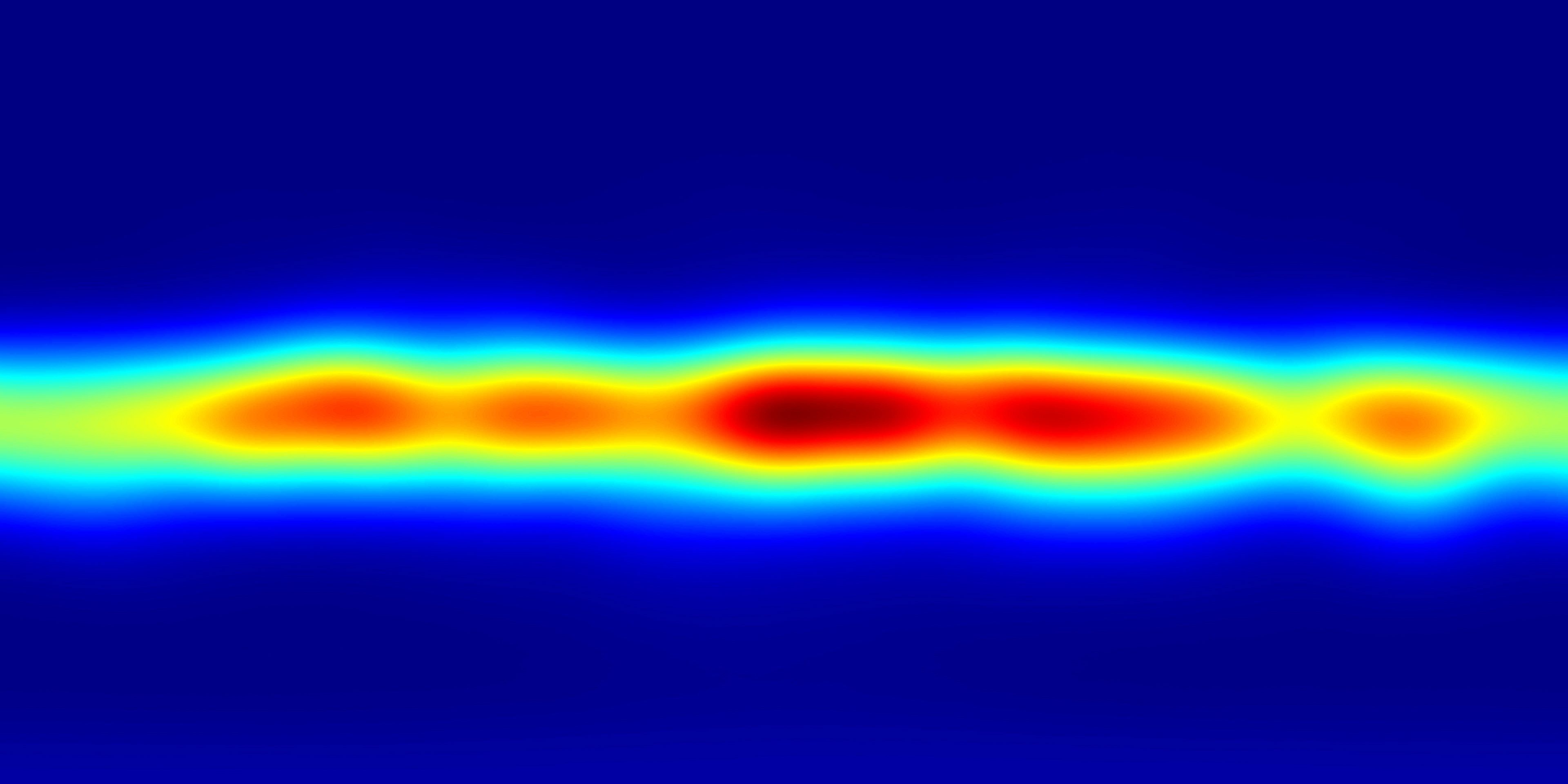}}\vspace{-12pt}
\caption{Overall heat maps of visual attention distribution for three audio modalities, respectively.} \label{fig:overlay-all}
\vspace{-8pt}
\end{figure}

\vspace{-13pt}
\subsubsection{Procedure.}
The complete subjective experimental process includes practice with the HMD, eye-tracking calibration and formal eye-movement collection. Subjects were first asked to wear the HTC Vive Pro Eye and observe a video example to ensure they were comfortable with the virtual environment before the formal experiment. After completing the practice, an eye-tracking calibration procedure was performed to ensure the accuracy of eye tracking. The subjects were then told to freely observe the presented videos in the VR-HMD. The longitude of the viewing starting point for each video was fixed to the 180$^{\circ}$ longitude line of the ODV, \textit{i.e.} the vertical center line of the ERP format. Considering the total length of the video, to avoid visual fatigue that could affect experimental results, the 162 videos were divided into two parts, with breaks allowed between them. There was also a 2-second black screen between every 2 ODVs. The overall duration of the experiment was 50-60 minutes per person.

%\subsection{Gaze Data Processing}
%As the raw eye-tracking data are recorded in the format of coordinates $(x, y, z)$ of gaze points on a sphere of radius 1 in three-dimensional Cartesian coordinate system
%, we first convert them into longitude-latitude coordinates. We use a two-step spatial dispersion threshold method \cite{Krassanakis_Filippakopoulou_Nakos_2014} to filter out saccades which have little practical meaning of visual attention and therefore obtain fixation. We superimpose fixation of 20 subjects for every frame to gain binary fixation maps. Since human eyes actually focus on the $2^{\circ}\sim 4^{\circ}$ area around fixation \cite{GUTIERREZ201835}, a Gaussian filter with a $3.34^{\circ}$ visual angle of view is applied to obtain saliency maps \cite{A_Dataset_of_Head_and_Eye_Movements_for_360_Degree_Images}.
\vspace{-8pt}
\section{Gaze Data Analysis}\label{sec:gaze_analysis}
\vspace{-3pt}
\subsection{Overall Attention Differences Between Three Modalities}
Fig.~\ref{fig:overlay-all} shows the overall visual attention differences between the three audio modalities, which are generated via superimposing all visual attention maps of all 162 ODVs.
It can be observed that attention in ambisonics modality is most widely distributed in longitude, especially in the area outside the initial observation area.
The mono case covers slightly less longitude than ambisonics.
In the mute modality, the attention distribution in the area far away from the initial observation point is relatively less.
It can be inferred that the presence of audio can to some extent guide the observer to look at a wider area rather than focusing near initial observation point, and ambisonics modality has better guidance.

We further use three consistency measures including CC, NSS and KLD to calculate the consistency of visual attention between modalities. To correct the distortion of ERP format where the poles are overstretched, we use a spiral sampling method to generate uniformly sampled points on the sphere, then project these sampled points onto the ERP image, and finally perform the consistency calculation based on sampled saliency maps. %\cite{A_Dataset_of_Head_and_Eye_Movements_for_360_Degree_Images}. 
Furthermore, to avoid interference from the viewing starting point, we only average the consistency of frames after the first second of each video as the result.
The ``overall'' rows of Table~\ref{tb:3type consistency} shows the CC and NSS scores between mute modality and ambisonics modality are the lowest, and the KLD score is the highest, indicating the lowest consistency of attention between these two modalities.
Mute modality and mono modality have the second lowest consistency, suggesting that the presence of audio does have an effect on the distribution of attention, and that ambisonics audio has a stronger effect 
compared to mono audio. 
Mono modality and ambisonics modality have the highest consistency, suggesting that there is less difference in the distribution of attention between two modalities.

%To quantify the general differences, we use three consistency measures to calculate the consistency of visual attention between modalities, namely Pearson's Correlation Coefficient (CC), Normalized Scanpath Saliency (NSS) \cite{PETERS20052397} and Kullback-Leibler Divergence (KLD).

%Table~\ref{tb:general consistency} shows the CC and NSS scores between mute modality and ambisonics modality are the lowest, and the KLD score is the highest, indicating the lowest consistency of attention between these two modalities. Mute modality and mono modality 
%Mute modality and mono have the second lowest consistency, suggesting that the presence of audio does have an effect on the distribution of attention, and that ambisonics audio has a greater effect 
%compared with mono audio. 
%Mono modality and ambisonics modality has the highest consistency, suggesting that there is less difference in the distribution of attention between two modalities.

\begin{figure}[!t]
\subfigure[Saliency maps of type 1 video at the 10th second]{\label{fig:type1-sal}
\begin{minipage}[b]{1\textwidth}
\centering
\includegraphics[width =1\textwidth]{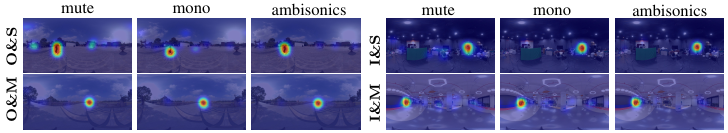}
\end{minipage}
}\vspace{-3pt}
\subfigure[Saliency maps of type 2 video at the 10th second]{\label{fig:type2-sal}
\begin{minipage}[b]{1\textwidth}
\centering
\includegraphics[width = 0.995\textwidth]{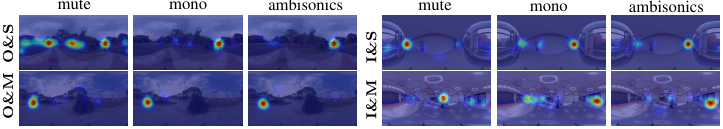}
\end{minipage}
}\vspace{-3pt}
\subfigure[Saliency maps of type 3 video at the 10th second]{\label{fig:type3-sal}
\begin{minipage}[b]{1\textwidth}
\centering
\includegraphics[width = 1\textwidth]{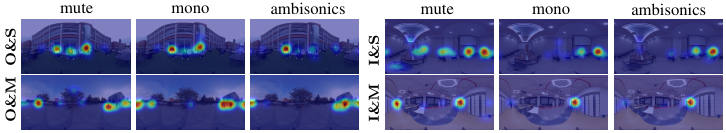}
\end{minipage}
}\vspace{-12pt}
\caption{Visual attention maps of 3 types of videos at the 10th second(\textit{i.e.} frame 300). For each triplet, the three columns are mute modality, mono modality, and ambisonics modality from the left to the right, respectively.
%The first column corresponds to mute modality, the second to mono modality and the third to ambisonics modality.
\label{fig:3type-sal}} \vspace{-6pt}
\end{figure}

\subsection{Influence of Various Audio Modalities on Various ODV Scenes}
Fig.~\ref{fig:3type-sal} shows several examples of the visual attention map of various audio modalities and various ODV scene categories at the 10$^{th}$ second (10-th second is relatively more discriminative).
It can be observed that the influence of audio on visual attention is more significant for scene type 2 compared to type 1 and 3.
Specifically, for type 1, observer's attention in mute modality is slightly more scattered.
For type 2, attention in both ambisonics modality and mono modality is mainly focused on the sound source, and the attention distribution is somewhat more concentrated in ambisonics modality, while the attention in mute modality is mainly distributed over visually salient objects.
For type 3, due to the complexity of salient objects and sound sources, the distribution of the observer's attention is more scattered compared to type 1 and type 2, and the mute modality is the most scattered. 
The distribution in mono modality is similar to that in ambisonics modality, and the ambisonics modality is more concentrated.

We also calculate the consistency between different modalities in three types of video and show the results in Table.~\ref{tb:3type consistency}, which further verifies our conclusions from Fig.~\ref{fig:3type-sal}.
For type 2 videos, the difference between mute modality and two sound modalities is the biggest, suggesting that sound has the greatest effect on visual attention when there are multiple visually salient objects and only one sound source, while sound has less effect on visual attention in the other two types. The smaller difference between the two sound modalities suggests that the orientation of sound does not have a significant effect on the observer's attention in the voiced situation.

\begin{table}[!t]
\caption{The consistency between different modalities in three types of video}\label{tb:3type consistency}\vspace{-5pt}
\centering
\scalebox{0.8}{
\begin{tabular}{|c|c|c|c|c|}
 \hline
\multirow{2}*{Metrics}& \multirow{2}*{Type} &\multicolumn{3}{c|}{Modality 1 - Modality 2} \\
\cline{3-5}
&	&	mute-mono&mute-ambisonics&mono-ambisonics\\
\hline
& 1&0.7859& 0.7677& 0.8398 \\
CC &2 &0.7646&0.7314& 0.8300\\
& 3&0.7823& 0.7558& 0.8264\\
& \,overall\, &0.7771&0.7508&0.8319\\
\hline
& 1 & 3.4176 & 3.4723 & 4.0730  \\
NSS & 2 &3.3080 & 3.2172 & 3.8508 \\
& 3 &3.3797 & 3.3093 & 3.8189\\
&overall&3.3660&3.3277&3.9103\\
\hline
& 1 &3.1402 &3.3909 &2.7259  \\
KLD & 2 &2.8249 & 3.1295 & 2.4767 \\
& 3 &2.6172 & 2.8370& 2.2343\\
&overall&2.8562&3.1161&2.4758\\
\hline
\end{tabular}}\vspace{-10pt}
\end{table}

\vspace{-12pt}
\begin{figure}[!t]
\centering
\includegraphics[width=0.9\textwidth]{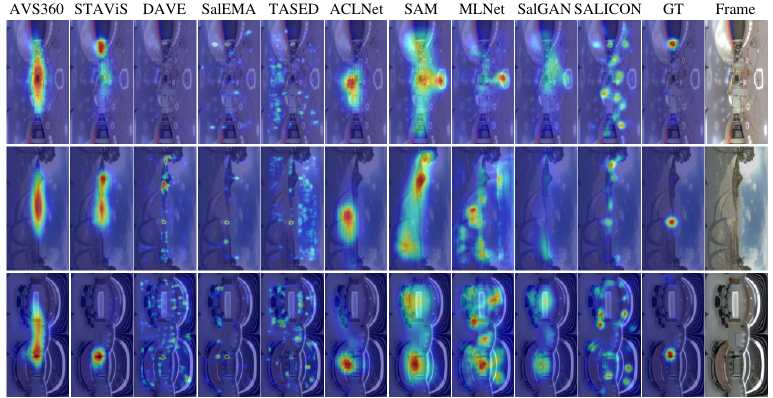}\vspace{-10pt}
\caption{Examples of results predicted using existing algorithms.}\vspace{-18pt} \label{fig:baseline}
\end{figure}

\vspace{-3pt}
\section{Baseline Performance}
\vspace{-8pt}
%%%%%%%%%记得加引用！！！！！！！！！！！！！
We test the performance of 10 state-of-the-art saliency models based on our AVS-ODV dataset, including four traditional image saliency prediction algorithms (SALICON~\cite{7410395}, MLNet~\cite{7900174}, SalGAN~\cite{Pan2017SalGANVS}, SAM~\cite{8400593}), three traditional video saliency prediction algorithms (TASED-Net~\cite{min2019tased}, ACLNet~\cite{8744328}, SalEMA~\cite{linardos2019simple}), two traditional video audiovisual saliency prediction algorithms (STAViS~\cite{tsiami2020stavis}, DAVE~\cite{tavakoli2019dave}), and an omnidirectional video audiovisual saliency prediction algorithm (AVS360~\cite{9301766}). Examples of the predicted saliency maps are shown in Fig.~\ref{fig:baseline}.

We adopt six commonly used metrics including NSS, SIM, CC, AUC-J, shuffled AUC and KLD to quantitatively compare the performance. To better predict the actual attention of users watching through VR-HMDs with ambisonics, we use the saliency maps of users in ambisonics modality as Ground Truth. Still, we correct the projection distortion during evaluation and eliminate the interference of the viewing starting point.

% To correct the distortion of ERP projection format where the poles are overstretched, we use spiral sampling method to generate uniformly sampled points on the sphere, then projected these sampled points onto the ERP image, and then performed the evaluation based on sampled saliency maps \cite{A_Dataset_of_Head_and_Eye_Movements_for_360_Degree_Images}.
It can be observed that SalGAN~\cite{Pan2017SalGANVS} achieves the best performance for almost all metrics, while SalEMA~\cite{linardos2019simple} and STAViS~\cite{tsiami2020stavis} perform better on NSS and KLD criteria, respectively.
% 图像显著性预测模型对小尺寸的显著对象预测效果普遍较差，其中预测相对准确的是 MLNet 模型，它基本能够预测出显著对象的位置，并且假阳性区域较小。视频显著性模型在运动对象的显著性预测上表现好一些，并且预测的显著性区域更加精确。其中，表现最好的是SalEMA模型，它在除了多显著物单声源的大多数情况下都能精确地输出显著对象的位置。音视频显著性模型整体的预测效果没有预想中的优秀，DAVE 模型在不同场景下的预测效果相差较大，AVS360 呈现过度的赤道中心偏置，STAViS 在多数情况下可以预测出显著对象或声源的方位，但具体的定位不是非常精确。
%Image saliency prediction models are generally poor at predicting salient objects of small size. The MLNet model is relatively more accurate, which can largely predict the location of salient objects and has a small false positive area. The video saliency models perform somewhat better in predicting the saliency of moving objects, and predict the saliency region more accurately. The best one is the SalEMA model, which accurately outputs the location of salient objects in most cases except for scenes with multiple salient objects and a single sound source. The overall predictions of the audio-video saliency models are not as good as expected, with the DAVE model varying considerably between scenes, AVS360 showing an excessive equatorial centre bias, and STAViS not being accurate enough in localising salient objects or sound sources.

\begin{table}[!t]
\vspace{10pt}
\caption{Evaluation of prediction results of existing
algorithms.}\label{tb:baseline-eval}\vspace{-5pt}
\setlength{\tabcolsep}{6pt}
\centering
\begin{tabular}{|l|l|l|l|l|l|l|}
\hline
Model         & NSS$\uparrow$            & SIM$\uparrow$             & CC$\uparrow$              & AUC-J$\uparrow$          & s-AUC$\uparrow$           & KLD$\downarrow$ \\
\hline
SALICON~\cite{7410395}   & 1.3852 & 0.3176 & 0.2871 & 0.7622 & 0.7367 & 6.5451  \\
MLNet~\cite{7900174}     & 1.5431 & 0.3585 & 0.3920 & 0.7765 & 0.7588 & 6.0986  \\
SalGAN~\cite{Pan2017SalGANVS}    & 1.4651 & \textbf{0.3762} & \textbf{0.3995} & \textbf{0.8285} & \textbf{0.7973} & 5.5109  \\
SAM~\cite{8400593}       & 1.3221 & 0.3679 & 0.3807 & 0.7939 & 0.7632 & 5.7149  \\
TASED-Net~\cite{min2019tased} & 1.9814 & 0.1901 & 0.0697 & 0.6834 & 0.6905 & 10.123 \\
ACLNet~\cite{8744328}    & 0.8398 & 0.3176 & 0.2444 & 0.7402 & 0.6579 & 5.9416  \\
SalEMA~\cite{linardos2019simple}    & \textbf{2.8371} & 0.3038 & 0.3099 & 0.7637 & 0.7006 & 7.2814  \\
STAViS~\cite{tsiami2020stavis}    & 1.3703 & 0.3735 & 0.3628 & 0.7542 & 0.7134 & \textbf{5.3603}  \\
DAVE~\cite{tavakoli2019dave}      & 1.5331 & 0.2447 & 0.1275 & 0.6967 & 0.6598 & 8.0272  \\
AVS360~\cite{9301766}    & 0.9944 & 0.3138 & 0.2645 & 0.6626 & 0.6676 & 7.7618 \\
\hline
\end{tabular}\vspace{-18pt}
\end{table}\vspace{-8pt}

\section{Conclusion}
\vspace{-2pt}
This work presents a comprehensive omnidirectional audio-visual saliency dataset AVS-ODV, which involves 162 8K-resolution omnidirectional videos with ambisonics and the eye-tracking data collected from 60 participants under three audio modalities. 
Qualitative and quantitative analyses are conducted to investigate the impact of different audio modalities on the visual attention of observers across different types of video scenes, and the analysis results demonstrate that audio information has a significant influence on viewing behavior. 
Moreover, ten state-of-the-art saliency prediction models are tested on the AVS-ODV dataset, and a new benchmark for the constructed AVS-ODV dataset is established. 
%Evaluating their performance, results indicate that SalEMA and STAViS are superior in saliency prediction among all models, with SalEMA being highly-appropriate for further training and testing due to its simple structure, fast runtime, and excellent performance. 
%These findings facilitate further design of saliency prediction models as a reference and performance evaluation in other  audio-visual saliency datasets.
Our AVS-ODV dataset, corresponding analysis, and established benchmark can facilitate the further design of corresponding saliency prediction models for ODVs with audios.

\vspace{-5pt}
\subsubsection* {Acknowledgements.} This work is supported by National Key R\&D Project of China (2021YFF0900503), NSFC (61831015, 62101325, 62101326, 62271312, 62225112), Shanghai Pujiang Program (22PJ1407400), Shanghai Municipal Science and Technology Major Project (2021SHZDZX0102), STCSM (22DZ2229005).

\vspace{-8pt}
\bibliographystyle{splncs04}
\bibliography{AVS}




% LNCS Homepage, \url{http://www.springer.com/lncs}. Last accessed 4
% \end{thebibliography}
\end{document}